# Automatic Speech Recognition of Non-Native Child Speech for Language Learning Applications

**Simone Wills** ✉
Radboud University Nijmegen, The Netherlands

**Yu Bai** ✉
Radboud University Nijmegen, The Netherlands
NovoLearning, The Netherlands

**Cristian Tejedor-García** ✉
Radboud University Nijmegen, The Netherlands

**Catia Cucchiarini** ✉
Radboud University Nijmegen, The Netherlands

**Helmer Strik** ✉
Radboud University Nijmegen, The Netherlands

## Abstract

Voicebots have provided a new avenue for supporting the development of language skills, particularly within the context of second language learning. Voicebots, though, have largely been geared towards native adult speakers. We sought to assess the performance of two state-of-the-art ASR systems, Wav2Vec2.0 and Whisper AI, with a view to developing a voicebot that can support children acquiring a foreign language. We evaluated their performance on read and extemporaneous speech of native and non-native Dutch children. We also investigated the utility of using ASR technology to provide insight into the children's pronunciation and fluency. The results show that recent, pre-trained ASR transformer-based models achieve acceptable performance from which detailed feedback on phoneme pronunciation quality can be extracted, despite the challenging nature of child and non-native speech.

**2012 ACM Subject Classification** Human-centered computing → Human computer interaction (HCI); Applied computing → Education

**Digital Object Identifier** 10.4230/OASIcs.SLATE.2023.11

**Category** Short Paper

**Funding** The project ST.CART is funded by the European Regional Development Fund (ERDF).

**Acknowledgements** Special thanks go to all the children who participated, their parents, their teachers, and the schools.

## 1 Introduction

In the field of education there is considerable interest in developing and employing voicebots to support children in fundamental skills like learning to read or in acquiring foreign languages. However, this is one of the areas in which the performance of automatic speech recognition (ASR) systems is still lacking. In general, current speech-based systems work relatively well for adult native speakers, with the performance dropping considerably when it comes to other target groups such as elderly speakers, children, non-native speakers, and patients with pathological speech [4, 8]. There are two main factors which contribute to this; firstly the inherently challenging nature of the the speech produced by these speakers and secondly, the limited availability of the speech data needed to develop and improve these systems.

In terms of child speech, ASR performance is hindered by speech characteristics which include high degrees of variability, linguistic differences as a result of different stages of language acquisition, and the marked presence of disfluencies and hesitation. This is in addition to physiological differences, such as shorter vocal tracts, which cause a mismatch between the adult and child speech [15]. For applications geared towards supporting the acquisition of a second language (L2) or foreign language by children, the task is doubly challenging in that it deals not only with child speech, but that on non-native speech which is equally as challenging for ASR systems [9]. These challenges are exacerbated by the limited availability of speech data, a particularly prominent problem when working with languages other than English [13]. Collecting speech data is also often hampered by the difficulty in obtaining approval for data collection as well as difficulty in gaining access to the speakers, as they are generally less independent.

Existing ASR-based educational applications built for adult L2 learners have demonstrated the utility of ASR output to provide immediate feedback on correct pronunciation based on single utterances [10, 16]. Despite the challenges, there are few educational ASR-based educational products which have been developed for children learning foreign languages. [11] built a system for children with pronunciation difficulties to practice in their native languages, where children get feedback, in the form of correct / incorrect, from an ASR system . In [17] English pronunciation of short utterances is scored based on tone, volume, timbre and speed.

In this paper we report on research conducted within the framework of a project aimed at investigating the usability of ASR technology for developing innovative educational applications for children learning a second language. In general, ASR technology works better when the text to be recognized is known in advance, as in the case of read speech. However, using only read speech considerably limits the possibilities for developing educational applications, which would indeed be confined to learning to read. For developing more creative applications, less prepared and more spontaneous speech would be required, but this makes the ASR even more complex [12].

To gain insight into the potential of ASR for these tasks in a language other than English, we conducted experiments in which we analyzed extemporaneous speech produced by non-native children learning Dutch through traditional and more recent ASR systems. The research questions we addressed are: *RQ1* How do current state-of-the art ASR systems fare on extemporaneous, Dutch non-native child speech, and *RQ2* how does their performance compare to that on native and read speech? We also consider what kind of information can be automatically extracted from such recordings to gain insights into language proficiency, such as measures of pronunciation quality, articulation and fluency.

The paper is organized as follows. Section 2 describes the data we used, while Section 3 describes the ASR system evaluations. We present both the analysis and ASR performance results in Section 4, followed by a discussion thereof in Section 5. The paper concludes with final remarks and possible future work in Section 6.

## 2    Speech Material

In this study, we use speech material taken from the JASMIN corpus [7]. This is a Dutch-Flemish corpus of around 90 hours of contemporary Dutch speech from groups of speakers not represented in the existing Spoken Dutch Corpus (Corpus Gesproken Nederlands; CGN). This includes children, non-native speakers, and elderly people. One of the corpus aims was to collect speech from human-machine interactions. The corpus includes manual orthographic transcriptions and phonetic transcriptions produced automatically using ASR.

For our study we selected the Dutch human-machine dialogues of children aged between 11 and 18 years old, separated into two groups; native Dutch child speakers and non-native children learning Dutch. This corresponds to component-p of Group 2 and 3 in the corpus structure. The total selected data contains 9 hours and 21 minutes of dual-channel speech recordings, produced by 94 children (1).

In these dialogues children are questioned about activities they enjoy and are stimulated to provide answers on the fly. The dialogues are conducted in a Wizard-of-Oz scenario, that is that the role of the computer is played by a human being in disguise and the answers provided by this person are sounded through TTS to make them sound like computer-generated speech. The resulting speech is considered extemporaneous speech, as opposed to spontaneous, because the children were prompted through questions to produce answers rather than generating speech on their own initiative.

These recordings were selected because they reflect the realistic challenges which are encountered when applying ASR to turn-taking conversations with children, as one might expect in language learning practice. Additionally, these dialogues were designed such that the children would produce hesitations and dysfluencies, as is often the case in extemporaneous human-machine communication.

**Table 1** Overview of Data

|            | No. Participants | Female | Male | Age (Years) | Duration (Hour) |
|------------|------------------|--------|------|-------------|-----------------|
| **Native**     | 41               | 21     | 20   | 12-18       | 4.48            |
| **Non-Native** | 53               | 28     | 25   | 11-18       | 4.87            |
| *Total*    | 94               | 49     | 45   | 11-18       | 9.35            |

## 3 Methodology

### 3.1 ASR Systems

We compare the performance of two state-of-the-art open-source ASR systems: Wav2Vec2.0 [1] and Whisper AI [14]. Both systems employ openly available pre-trained models . Wav2Vec2.0 is a transformer-based model pre-trained on unlabelled audio, which can then be further fine-tuned. In this paper we use a cross-lingual model[1], which has been trained on multiple languages and fine-tuned on Dutch. The Whisper ASR system is a recent release by OpenAI, and is an encoder-decoder transformer model pre-trained on multilingual data. Word-level time-stamps are obtained using the WhisperX Python library [3].

### 3.2 Speech Characteristics

We are not only interested in how well the different ASR systems perform in decoding the audio, but whether this output can be reliably used to automatically extract information about the children's language proficiency. We calculated a number of measurements, for each speaker, relating to speech characteristics. The measurements and their calculations are presented in Table 2.

---

[1] `https://huggingface.co/FremyCompany/xls-r-2b-nl-v2_lm-5gram-os`

**Table 2** Speech Characteristic Measurements

| Measurement | Calculation |
| --- | --- |
| Number of utterances | segments of speech surrounded by significant pause |
| Number of words | count of all word tokens |
| Number of phones | count of all phone tokens |
| Number of filled pauses | count of all filled pause words (e.g. "hmm", "uh", "ehm") |
| Vocab Size | count of unique word tokens |
| Average number of words per utterance | number of words / number of utterances |
| Average number of phones per utterance | number of phones / number of utterances |
| Total word duration (seconds) | sum of all word durations |
| Average word duration (seconds) | total word duration / total number of words |
| Articulation rate | number of phones / total word duration |
| Total speech duration (seconds) | sum of all utterance durations |

We first calculated these measures through manual transcriptions and then used the Whisper output to calculate a subset of these measures, so that we can compare manual transcription-based measures to those based on ASR output. For the Whisper output, the text was also lower-cased and punctuation removed for measurements replying upon word forms, such as vocab size and filled pause identification.

## 3.3 Pronunciation Evaluation

Pronunciation information was extracted using the NovoLearning ASR system. This is a back-end ASR system in a Dutch automatic Reading Tutor application which "listens" and provides feedback to children. The ASR takes prompts and speech recordings as input and gives results on both word level and phone level. The ASR analyses speech and gives confidence score represented by probabilities on each phone of the words in the prompt [2]. The probability score ranges from 0 to 100. We passed the native and non-native dialogue speech with the manual transcripts to the NovoLearning ASR. Unrecognized words and filler words were filtered out. We calculated the mean confidence score and standard deviation (SD) of each phoneme for native and non-native speech. We used the phone probability scores to represent the pronunciation quality of each phoneme.

## 4 Results

## 4.1 ASR Performance

The word error rate (WER) for Whisper versus Wav2Vec2.0 on the native and non-native speech are shown in Table 3. WER measures the accuracy of the text output of an ASR system by computing the percentage of words which it transcribed incorrectly, determined through comparison with a 'gold standard' (typically manual) transcription. A higher WER indicates more inaccuracies. As seen 3 Whisper outperforms Wav2Vec2.0, so we further compared the performance of Whisper on Read vs. Dialogue speech. The WERs for Read speech are lower than those for Dialogue, with the WER for native Read speech being particularly low compared to the others.

**Table 3** WERs (%) using two ASR models for native and non-native Dialogue Speech

| Model | Native | Non-native |
|---|---|---|
| Wav2Vec | 44.80 | 44.60 |
| Whisper | 30.70 | 33.80 |

**Table 4** WER (%) of Whisper for Read vs. Dialogue Speech

|  | Read | Dialogue |
|---|---|---|
| Native | 8.00 | 30.70 |
| Non-native | 24.80 | 33.80 |

## 4.2 Speech Characteristics

**Table 5** Speech Characteristics: Median and InterQuartile Range (IQR) values for manual transcriptions

| Speech Characteristic | Native | | Non-native | |
|---|---|---|---|---|
|  | Median | IQR | Median | IQR |
| Number of utterances | 40 | 10 | 52 | 18 |
| Number of words | 105 | 60 | 141 | 84 |
| Number of phones | 375 | 147 | 482 | 251 |
| Number of filled pauses | 3 | 7 | 14 | 12 |
| Vocab size | 68 | 37 | 84 | 34 |
| Average words / utterance | 2.51 | 1.35 | 2.82 | 0.66 |
| Average phones / utterance | 9.39 | 3.99 | 9.44 | 1.90 |
| Total word duration | 39.97 | 13.31 | 53.40 | 31.31 |
| Average word duration | 0.38 | 0.22 | 0.38 | 0.37 |
| Articulation rate | 9.59 | 1.48 | 8.94 | 1.54 |
| Total speech duration | 48.93 | 15.79 | 68.94 | 29.96 |

**Table 6** Speech Characteristics: Median and InterQuartile Range (IQR) values for Whisper output.

| Speech Characteristic | Native | | Non-native | |
|---|---|---|---|---|
|  | Median | IQR | Median | IQR |
| Number of words | 220 | 68.5 | 262 | 57 |
| Number of filled pauses | 1 | 1.5 | 3 | 4 |
| Vocab size | 129 | 20.5 | 145 | 24 |
| Average words / utterance | 4.25 | 0.70 | 4.17 | 0.85 |
| Total word duration | 57.95 | 22.95 | 72.66 | 13.30 |
| Average word duration | 0.26 | 0.04 | 0.28 | 0.02 |

Table 5 and Table 5 present the speech characteristic measurements we calculated based on the manual and Whisper ASR-based transcriptions, respectively. Using the manual transcriptions, we see that for several measures, in comparison to native-speakers, the non-native speakers have higher median values (the number of utterances, words, phones, and filled pauses, total speech duration, total word duration, and vocabulary size). Articulation rate has a lower value, and average number of words and phones per utterance have similar values in the two groups. A subset of measurements calculated using Whisper output are given in Table 6. While the values are all higher when compared to the manual-based values, but the same trends are reflected in the data, with non-native speaker values being higher than native speakers' for the same characteristics as above.

## 4.3  Pronunciation Evaluation

**Table 7** Mean and SD values for the phoneme confidence scores for native and non-native speech.

|     | Native | | Non-native | |     | Native | | Non-native | |
| --- | --- | --- | --- | --- | --- | --- | --- | --- | --- |
|     | Mean | SD | Mean | SD |     | Mean | SD | Mean | SD |
| a   | 95.69 | 12.85 | 92.32 | 15.61 | b   | 97.07 | 8.38  | 96.65 | 8.87  |
| aa  | 96.97 | 11.26 | 96.32 | 11.59 | d   | 97.51 | 8.94  | 97.10 | 10.27 |
| aw  | 95.12 | 11.67 | 88.13 | 18.29 | f   | 96.30 | 7.55  | 93.83 | 12.47 |
| ax  | 94.60 | 15.16 | 95.21 | 12.91 | g   | 34.18 | 27.08 | 44.99 | 28.36 |
| eh  | 90.99 | 19.97 | 90.58 | 19.27 | hh  | 75.94 | 35.26 | 76.32 | 32.41 |
| ei  | 96.73 | 10.84 | 96.79 | 8.83  | k   | 98.23 | 7.26  | 97.52 | 9.43  |
| eu  | 98.42 | 4.59  | 85.25 | 24.39 | l   | 97.84 | 8.06  | 96.52 | 9.87  |
| ey  | 90.73 | 22.71 | 90.75 | 21.43 | m   | 97.27 | 8.10  | 96.18 | 9.70  |
| ih  | 94.84 | 12.30 | 92.79 | 15.21 | n   | 97.48 | 8.57  | 97.56 | 7.99  |
| iy  | 98.12 | 6.14  | 92.70 | 15.72 | ng  | 97.78 | 9.52  | 96.37 | 10.58 |
| oh  | 92.08 | 17.38 | 87.04 | 19.91 | p   | 98.05 | 7.15  | 97.48 | 9.30  |
| ow  | 96.80 | 10.24 | 93.65 | 14.46 | r   | 97.20 | 9.11  | 95.29 | 12.31 |
| uh  | 93.28 | 12.72 | 86.64 | 19.77 | s   | 98.74 | 4.51  | 96.57 | 11.20 |
| uu  | 86.91 | 20.89 | 72.64 | 27.91 | sh  | 95.67 | 8.43  | 72.16 | 28.79 |
| uw  | 84.57 | 29.93 | 77.13 | 31.42 | t   | 97.14 | 10.26 | 96.29 | 12.71 |
| uy  | 97.31 | 6.64  | 87.25 | 19.10 | v   | 98.43 | 6.17  | 97.40 | 9.94  |
|     |       |       |       |       | wv  | 94.68 | 15.63 | 93.68 | 15.52 |
|     |       |       |       |       | x   | 97.90 | 8.26  | 97.75 | 8.63  |
|     |       |       |       |       | y   | 89.42 | 25.05 | 85.99 | 26.53 |
|     |       |       |       |       | z   | 98.28 | 5.31  | 95.66 | 12.75 |
|     |       |       |       |       | zh  | 96.24 | NaN   | 70.87 | 40.07 |

Table 7 shows that the mean probability scores of most phonemes are similar between native and non-native speech. Most phonemes are above 90. The probability of phoneme "g" is low in both groups (34.18 for native speech, and 44.99 for non-native speech). We looked into the words containing the "g" sound, most of them are English words since the "g" sound does not occur in Dutch words. The only two Dutch words containing the "g" sound are words "stokbrood", "honkbal" and its plural form "honkballen". The "g" sound in these words is a result of voice assimilation. The voiceless velar "k" changes to a voiced velar stop. However, the velar stop "g" is different from the English "g", which leads to low probability. the confidence scores for "hh" and "y" are slightly lower than those of other consonants in both groups. For native speech we observe slightly higher scores in vowels like "eu", "uu" and "uy". However, score differences for these vowels between the two groups are still below 20.

## 5 Discussion

In the present study we investigated the performance of the latest state-of-the-art ASR systems on Dutch child speech to determine to what extent this technology can be employed to process extemporaneous speech produced by non-native children learning Dutch. As to our research question on how these ASR systems fare on extemporaneous, Dutch non-native child speech ($RQ1$), the results indicate that recent, pre-trained transformer-based models like Wav2Vec2.0 and Whisper obtain reasonable performance. Of these two transformer-based models, Whisper achieves the better performance. While not further explored in this paper, Whisper also has the advantage of producing punctuation. As to our second research question ($RQ2$) regarding the performance difference between native and non-native, read and extemporaneous speech, the results show that ASR performance is better for native than non-native speech, as expected. Similarly, better results were observed for read speech over extemporaneous speech. In particular, the results for read speech appear to be much better for the group of native speakers. There are two possible explanations for this finding. As native speakers, these children can read much better in Dutch. In addition, their realizations of the Dutch sounds are more accurate, which leads to better ASR performance.

To gain a deeper understanding of these differences, we extracted information about several aspects of speech quality that provide insights into language proficiency. We first analyzed differences in temporal measures such as articulation rate, duration and fluency and vocabulary-based measures such as number of words and vocabulary size. Temporal and fluency measures indicate that native speakers are much more fluent than non-native speakers. They speak faster and produce fewer dysfluencies, which is in line with previous research findings [6]. With respect to the amount of speech produced, the results show that the non-native speakers are surprisingly more talkative. They seem to produce longer replies to the questions posed by the computer, which contain more words, but these words appear to be shorter and often interrupted.

For pronunciation quality, however, we did not see considerable differences between native and non-native speakers. ASR-based confidence scores for individual speech sounds do not seem to differ between the two groups of native and non-native speakers, which is surprising.

It is clear that further research is needed to gain more insights into the differences between native and non-native speech and the way in which these affect ASR performance. Nevertheless, is is interesting to see that many of the results obtained through manual transcriptions do not differ considerably from those based on ASR output. This means that this technology has the potential to facilitate future analyses of non-native speech in the context of language learning and language evaluation as well as the design and development of innovative language learning applications.

## 6 Conclusions

The results of the present study on the suitability of state-of-the-art ASR technology for Dutch non-native read and extemporaneous child speech allow us to conclude that in spite of the considerable challenges recent, transformer-based models open up new perspectives for applying this technology in educational applications. The use of ASR technology also makes it possible to extract additional measures that provide insights into language proficiency. As has been shown previously [5] completely correct performance is not necessarily required to be able to realize pedagogically sound language learning applications.